\documentclass[conference]{IEEEtran}
\IEEEoverridecommandlockouts
\usepackage[table,xcdraw]{xcolor} 
\usepackage{float}      % 【关键！】提供 [H] 参数，强制固定图片位置

\usepackage{amsmath,amssymb,amsfonts}
\usepackage{textcomp}
\usepackage{url}
\usepackage{graphicx}
\usepackage{booktabs}   % 用于 \toprule, \midrule, \bottomrule
\usepackage{tabularx}   % 用于 X 列自动换行
\usepackage{multirow}   % 用于合并行
\usepackage{algorithm}
\usepackage{algorithmicx}
\usepackage{algpseudocode}
\usepackage{listings}

\usepackage{cite}
\usepackage{enumitem}   % 用于自定义列表样式
\usepackage[most]{tcolorbox}
\tcbuselibrary{skins, breakable}

\usepackage{times}      % 使用 Times 字体
\usepackage{inconsolata}% 使用 Inconsolata 字体（可选）
\usepackage{microtype}  % 提高排版质量
\usepackage[utf8]{inputenc} % 支持 UTF-8 编码
\usepackage[T1]{fontenc}    % 更好的字符编码支持

\definecolor{bestorange}{HTML}{FFD580}
\definecolor{secondblue}{HTML}{D1EEFB}
\definecolor{customgreen}{HTML}{2E8B57}
\definecolor{forestgreen}{rgb}{0.0, 0.6, 0.0}

\def\BibTeX{{\rm B\kern-.05em{\sc i\kern-.025em b}\kern-.08em
    T\kern-.1667em\lower.7ex\hbox{E}\kern-.125emX}}
\makeatletter
\def\ps@IEEEtitlepagestyle{%
  \def\@oddfoot{\mycopyrighttxt}%
  \def\@evenfoot{}%
}
\def\mycopyrighttxt{%
  {\footnotesize \begin{minipage}{\textwidth}
  \linespread{1.0}\selectfont
  \copyright~2026 IEEE. Personal use of this material is permitted. Permission from IEEE must be obtained for all other uses, in any current or future media, including reprinting/republishing this material for advertising or promotional purposes, creating new collective works, for resale or redistribution to servers or lists, or reuse of any copyrighted component of this work in other works.
  \end{minipage}}%
}
\makeatother
\begin{document}
\bstctlcite{IEEEexample:BSTcontrol}
\title{Strat-LLM: Stratified Strategy Alignment for LLM-based Stock Trading with Real-time Multi-Source Signals\\
}
\author{
\IEEEauthorblockN{1\textsuperscript{st} Wenliang Huang$^{*}$}
\IEEEauthorblockA{\textit{School of Economics} \\
\textit{Zhejiang University of Technology}\\
Hangzhou, China \\
302023065053@zjut.edu.cn}
\and
\IEEEauthorblockN{2\textsuperscript{nd} Zengyi Yu}
\IEEEauthorblockA{\textit{Faculty of Education} \\
\textit{East China Normal University}\\
Shanghai, China \\
51284118014@stu.ecnu.edu.cn}
}
\maketitle

\begin{abstract}
Large Language Models (LLMs) are evolving into autonomous trading agents, yet existing benchmarks often overlook the interplay between architectural reasoning and strategy consistency. We propose \textbf{Strat-LLM}, a framework grounded in Stratified Strategy Alignment. Operating in a \textit{live-forward} setting throughout 2025, it integrates heterogeneous data including sequential prices, real-time news, and annual reports to eliminate look-ahead bias. Extensive stress tests on A-share and U.S. markets reveal: (1) reasoning-heavy models achieve peak utility in Free Mode via internal logic, whereas standard models require Strict Mode as a vital risk anchor; (2) alignment utility is regime-dependent, with Free and Guided modes capturing momentum in uptrending markets, while Strict Mode mitigates drawdowns in downtrends; (3) mid-scale models (35B) show optimal fidelity under strict constraints, whereas ultra-large models (122B) suffer an alignment tax under rigid rules but gain a performance premium in Guided Mode; (4) standard LLMs often fall into a high win-rate trap, optimizing for small gains at the expense of total returns, which can only be mitigated through deep reasoning or strict external guardrails. Project details are available at \url{https://Strat-LLM.github.io}.
\end{abstract}


\section{Introduction}

In recent years, large language models (LLMs) have transcended the paradigm of pure text generation, evolving into \textbf{autonomous agents} endowed with Multi-Source perception and complex decision-making capabilities. The very recent emergence of local-resident agents like \textbf{OpenClaw} confirms that LLMs can now technically function as persistent, autonomous operators rather than mere chatbots. However, this operational capacity merely exposes a critical knowledge gap: we have yet to discern specifically where their decision-making capabilities excel or falter when subjected to varying model parameters, trading horizons, risk attitudes, and distinct strategic execution modes (ranging from strict adherence to full autonomy). Within this paradigm shift, \textbf{financial trading}, owing to its direct linkage to economic value and its stringent requirements on decision logic, has emerged as a \textit{litmus test} for evaluating agent competence.

Existing evaluation frameworks generally fall into two categories. The first focuses on static knowledge, exemplified by benchmarks such as \textbf{FinBen}~\cite{xie2024finben} and \textbf{Pixiu}~\cite{xie2023pixiu}, which assess fundamental financial literacy. The second comprises dynamic environments like \textbf{FinTSB}~\cite{hu2025fintsb}, \textbf{StockBench}~\cite{chen2025stockbench}, which simulate trading behaviors. Recently, Multi-Source datasets such as \textbf{FinMME}~\cite{luo2025finmme} and \textbf{Time-MMD}~\cite{liu2024time} have begun to address heterogeneous data integration.

Despite this progress, three critical limitations persist. First is the absence of \textbf{strategic executability}. Most benchmarks emphasize final profits while neglecting whether the decision process adheres to predefined risk rules. As Wallace et al.~\cite{wallace2024instruction} argue, LLMs inherently struggle to prioritize privileged system instructions such as risk limits over conflicting lower-priority inputs like market noise, rendering them operationally dangerous. Second, existing evaluations overlook agents' \textbf{vulnerability to behavioral biases}. Standard frameworks fail to audit the structural soundness of an agent's trading behavior. Without rigid external constraints or intrinsic deep reasoning, LLMs frequently fall into a ``High Win-Rate Trap'', exhibiting a retail-like Disposition Effect where they optimize for frequent small gains but tolerate massive drawdowns during volatile regimes. Third, \textbf{data contamination} remains unresolved, as models may simply memorize historical patterns rather than reason dynamically through temporal shifts~\cite{lopez2025memorization}.

To address these challenges, we propose \textbf{Strat-LLM}, a \textbf{stratified strategy alignment framework} designed for the real financial market of 2025. We define three progressively constrained modes: \textbf{Free}, \textbf{Guided}, and \textbf{Strict} to explore the boundaries between intrinsic intuition and compliance. To audit strategic consistency, we require models to operate under explicitly predefined execution protocols that isolate the effect of rule adherence from autonomous reasoning. Furthermore, we employ a \textbf{T+1 rolling-window backtest} using real-time 2025 data to eliminate look-ahead bias. Our extensive evaluation reveals a fundamental \textbf{Reasoning Dichotomy} in how models respond to strategic constraints, and identifies an \textbf{Alignment Tax} where rigid rule enforcement incurs measurable performance costs that vary systematically with model scale and market regime.

\section{Related Work}

\subsection{Evolution of Financial Agents \& Dynamic Evaluation}
The evaluation paradigm for Financial Large Language Models (FinLLMs) has shifted from static text analysis to dynamic trading simulations. Early works~\cite{xu2018stock} established the baseline for predicting stock movements from textual signals. This evolved into static knowledge benchmarks like \textbf{FinBen}~\cite{xie2024finben} and \textbf{PIXIU}~\cite{xie2023pixiu}, which tested fundamental literacy but lacked temporal complexity.

To capture real-world market dynamics, recent frameworks adopted rolling-window backtesting. \textbf{FinTSB}~\cite{hu2025fintsb} and \textbf{StockBench}~\cite{chen2025stockbench} introduced environments where agents adapt to shifting market regimes. The push for automation is further exemplified by \textbf{KRX Bench}~\cite{son2024krx}, which automates financial benchmark creation, and \textbf{InvestorBench}~\cite{li2025investorbench}, which assesses financial decision-making tasks for agents. However, these dynamic evaluations prioritize outcome-oriented metrics (e.g., annualized return) over \textbf{strategic executability}, often failing to distinguish whether a model is reasoning financially or merely exploiting probability patterns~\cite{lopez2025memorization}.

\subsection{Multi-Source Decision-Making}
Financial trading is a process of conflict resolution across heterogeneous, multi-source information streams. While recent advances in multi-modal proxy learning and subspace clustering \cite{yao2024multi, yao2024customized} have improved the foundational representation of diverse data structures, applying these paradigms to dynamic finance remains challenging. Datasets like \textbf{FinMME}~\cite{luo2025finmme} and \textbf{Time-MMD}~\cite{liu2024time} have advanced the integration of chart and text analysis but often treat Multi-Source reasoning as a static task. In contrast, real-world markets present conflicting signals (e.g., positive earnings reports amidst technical breakdowns). \textbf{FinTMMBench}~\cite{zhu2025fintmmbench} benchmarks temporal-aware Multi-Source retrieval-augmented generation (RAG), highlighting the difficulty agents face in synthesizing heterogeneous data streams without succumbing to information overload.

\subsection{Strategic Alignment and Behavioral Biases}
Financial trading demands strict adherence to risk management protocols, a requirement that poses a significant challenge for current autonomous agents. Research shows that LLMs inherently struggle with instruction hierarchy; as Wallace et al.~\cite{wallace2024instruction} demonstrate, models often fail to prioritize privileged system instructions (e.g., rigid risk limits or strategy rules) when confronted with conflicting lower-priority inputs (e.g., volatile market data and news). In dynamic financial environments, this alignment failure frequently manifests as emergent behavioral biases. Instead of optimizing for mathematical expectancy, unconstrained standard models often default to retail-like heuristics, such as the Disposition Effect, prematurely realizing small gains to inflate win rates while failing to execute stop-losses effectively. Strat-LLM addresses this gap by shifting the evaluation paradigm from static profitability to dynamic behavioral auditability. By mandating stratified interaction modes (Free, Guided, Strict), our framework explicitly measures an agent's ability to overcome intrinsic cognitive biases, balance intuitive reasoning with rule adherence, and maintain strategic consistency across shifting market regimes.

\section{METHODOLOGY}
\begin{figure*}[t!]
    \centering
    \includegraphics[width=0.95\textwidth]{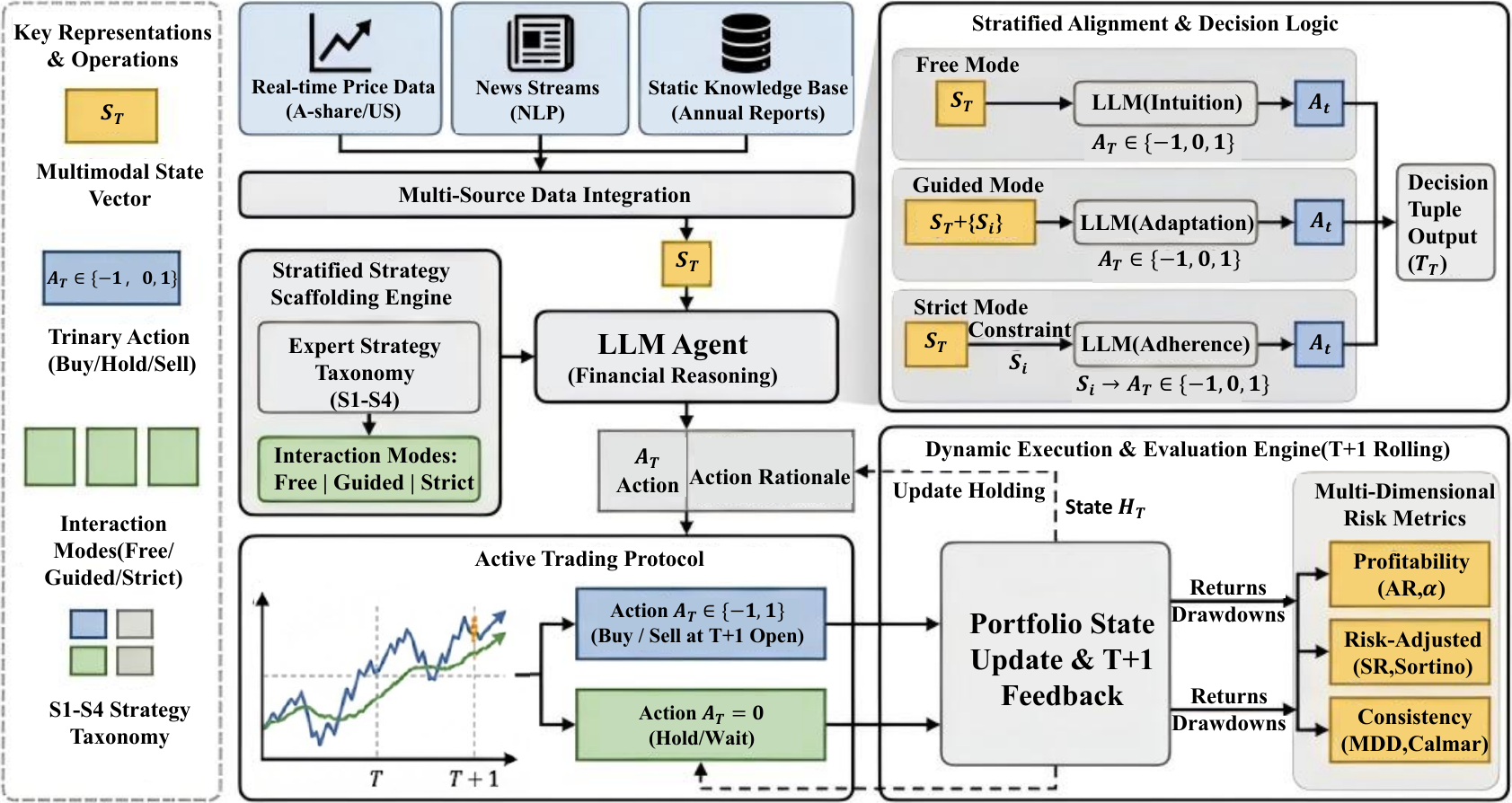} 
    \caption{\textbf{The overall architecture of Strat-LLM.} The framework operates in three sequential phases: (1) \textit{Multi-Source Data Integration}, which fuses real-time price data, news streams, and static knowledge into a unified state vector $S_T$; (2) the \textit{Stratified Strategy Scaffolding Engine}, which directs the LLM to generate trading actions $A_T$ under three varying levels of autonomy (Free, Guided, and Strict) based on the Expert Strategy Taxonomy (S1--S4); and (3) the \textit{Dynamic Execution \& Evaluation Engine}, which executes orders using a T+1 rolling-window mechanism and computes multi-dimensional risk metrics (e.g., Sharpe Ratio, Max Drawdown) to close the feedback loop.}
    \label{fig:framework}
\end{figure*}
We propose \textbf{Strat-LLM}, a comprehensive framework designed to audit the strategic alignment of Large Language Models (LLMs) in financial trading. The system simulates a realistic institutional trading environment and consists of three tightly coupled modules: (1) a \textit{Multi-Source Data Integration Module} that processes heterogeneous market signals and static knowledge; (2) a \textit{Stratified Strategy Scaffolding Engine} responsible for dynamically injecting varying levels of strategy constraints; and (3) a \textit{Dynamic Execution \& Evaluation Engine} that executes trades via a rolling-window mechanism and computes multi-dimensional risk metrics.

\subsection{Multi-Source Data Integration}
To construct a \textit{contamination-free} evaluation environment, we integrate heterogeneous data sources covering two major markets: A-Shares and U.S. Stocks.

\subsubsection{Data Construction \& Time Horizons}
The system was deployed in a \textit{live-forward} setting throughout the 2025 experimental window, utilizing robust knowledge fusion paradigms \cite{zhou2026dual} to ensure interpretable state representation across the multi-source streams. Unlike retrospective backtesting, the agent received data streams sequentially in real-time, strictly preserving the chronological flow of information to eliminate look-ahead bias. All decisions were executed based on the information available at the specific timestamp of the 2025 trading sessions. To accommodate different trading calendars and ensure data integrity, we set staggered evaluation intervals:
\begin{itemize}
    \item \textbf{U.S. Market:} Evaluation period from January 1, 2025, to June 30, 2025.
    \item \textbf{A-Share Market:} Evaluation period from June 1, 2025, to September 30, 2025.
\end{itemize}

Based on this, we designed a dual time-scale to comprehensively examine agent adaptability:

\paragraph{Short-term Tactical Windows}
We selected 3 independent short-term windows within each market's evaluation period, each containing approximately 15 trading days. For U.S. stocks, these windows are distributed from early to mid-2025, covering high-volatility periods post-earnings releases. For A-shares, windows focus on the oscillation and trend transition periods from June to September. This design tests the model's rapid reaction capabilities and tactical flexibility in the face of sudden microstructural changes (e.g., policy announcements, breaking news).

\paragraph{Long-term Strategic Window}
To evaluate capital management and strategic consistency over a long cycle, we set 1 continuous long-term window per market, covering approximately 90 trading days. The U.S. window covers January to May, experiencing a complete semi-annual market cycle. The A-share window covers June to late September. This requires the model to navigate a full market volatility cycle (oscillation, rally, or correction), verifying its ability to avoid ``Strategy Drift'' during long-term holding.

\paragraph{Data Hierarchy}
The system integrates three data streams: (1) Minute-level price data; (2) Real-time news streams (sentiment and key events extracted via NLP); (3) Static Expert Knowledge Base (structured annual reports processed via hierarchical summarization, containing only information public as of trading day $T$).

\subsection{Stratified Strategy Alignment Protocol}
To decouple the model's ``reasoning capability'' from its ``instruction following capability,'' Strat-LLM introduces a core stratified strategy scaffolding mechanism.

\subsubsection{Expert Strategy Taxonomy}
We formally define four classic quantitative trading strategies as evaluation baselines:
\begin{itemize}
    \item \textbf{S1 (Short-Term Reversal):} Based on the overreaction hypothesis in behavioral finance, capturing mean reversion opportunities after short-term plunges.
    \item \textbf{S2 (Breakout Momentum):} A trend-following strategy that triggers a buy signal when the price breaks the 3-day high.
    \item \textbf{S3 (Volatility Compression):} A volatility regime-switching strategy identifying accumulation patterns in low-volatility zones.
    \item \textbf{S4 (Price-Volume Confirmation):} Based on price-volume theory, validating the effectiveness of price upward trends.
\end{itemize}

\subsubsection{Interaction Modes}
We implement three autonomy-progressive interaction modes via prompt engineering:
\begin{itemize}
    \item \textbf{Free Mode:} Simulates a zero-shot reasoning scenario, where the model relies on its native financial intuition, but this can lead to unstable decision-making as it has no external strategy constraints to guide it.
    \item \textbf{Guided Mode:} Simulates a human-AI collaboration scenario where strategies S1-S4 are provided as reference, allowing the model to adjust them based on real-time news. However, the model may still face internal cognitive conflicts, leading to behavioral inconsistency when the externally provided strategies don't fully align with its pre-trained intuition.
    \item \textbf{Strict Mode:} Simulates a compliance risk control scenario, where the model must strictly adhere to predefined strategies (S1-S4). While this enhances stability and reduces overtrading, it can also induce "Cognitive Dissonance," causing the model's internal intuition to clash with rigid strategy rules. This cognitive dissonance often manifests as execution paralysis or suboptimal trade timing, allowing us to explicitly audit the "Alignment Tax."
\end{itemize}

\subsubsection{T+1 Rolling Decision \& Feedback}
The system employs day-by-day simulation logic:
\begin{itemize}
    \item \textbf{Input:} On trading day $T$, the agent receives the Multi-Source state vector $S_T$.
    \item \textbf{Decision:} The model outputs a structured instruction containing the action signal $Action \in \{-1,0,1\}$, rationale. $Action=1$ indicates Buy, $Action=-1$ indicates Sell, and $Action=0$ indicates Hold. In ``Strict Mode,'' the rationale must explicitly cite specific clauses from the strategy library.
    \item \textbf{Execution:} 
        \begin{itemize}
            \item If $Action=1$ and cash is sufficient, buy at the opening price of $T+1$ (deducting costs). If cash is insufficient, execute ``Cash Truncation'' (buy the maximum affordable shares).
            \item If $Action=-1$, sell the model-designated volume of shares at the opening price of $T+1$ (deducting costs).
            \item If $Action=0$, maintain the current position.
        \end{itemize}
    \item \textbf{Update:} After market close, update unrealized PnL and feed the new holding state $H_T$ to the next day's prompt.
\end{itemize}

\subsubsection{Evaluation Metrics}
We employ a multi-dimensional set of metrics to comprehensively evaluate the agent's trading performance. These metrics are categorized into three primary dimensions:
\begin{itemize}
    \item \textbf{Profitability:} Measured by Annualized Return (AR) and Alpha ($\alpha$), which capture the absolute and excess returns generated by the model's trading strategy.
    \item \textbf{Risk-Adjusted Returns:} Evaluated using the Sharpe Ratio and Sortino Ratio, providing insights into the returns earned per unit of total risk and downside risk, respectively.
    \item \textbf{Strategic Consistency:} Assessed via Maximum Drawdown (MDD) and the Calmar Ratio, which quantify the portfolio's resilience and risk-return profile during market downturns.
\end{itemize}

\section{EXPERIMENTS}
\subsection{Experimental Settings}

\subsubsection{Models and Baselines}
To establish a comprehensive evaluation of large language models (LLMs) in financial decision-making, we select a diverse suite of state-of-the-art models, ranging from prominent open-source weights to leading proprietary architectures. 
\begin{itemize}
    \item \textbf{Open-Source Frontier Models:} We incorporate an array of highly capable open models, including Qwen3.5-Plus, Qwen3.5-122B-A10B, Qwen3.5-35B-A3B, Qwen3.5-9B, DeepSeek-V3.2, GLM-5, Kimi-k2.5, and Minimax-m2.5. This selection allows us to thoroughly assess the baseline trading acumen of accessible, top-tier general-purpose LLMs.
    \item \textbf{Proprietary State-of-the-Art Models:} To gauge the approximate upper bound of current artificial intelligence capabilities within our benchmark, we evaluate leading closed-source systems such as GPT-5.4, Claude-4.5-Sonnet, and Gemini-3.1-Pro.
\end{itemize}
To guarantee a fair and rigorous comparison, all evaluated models operate under a unified testing framework, utilizing standardized efficient sampling and generation configurations~\cite{yao2024swift}. This entails identical broker configurations, standardized interaction interfaces, and consistent input modalities across all experimental conditions.

\subsubsection{Data and Evaluation Protocol}
All experiments are executed under an \textbf{Active Trading Protocol}, allowing models to dynamically issue both \textit{buy} and \textit{sell} orders based on changing market conditions. The primary experimental configurations are summarized as follows.

\paragraph{Markets and Assets.}
Our evaluation spans the \textbf{A-share} and \textbf{U.S. equity markets}. To enhance generalizability and mitigate idiosyncratic risks, we select representative equities across diverse sectors, including technology, manufacturing, and consumer goods. Performance metrics are then aggregated and averaged across these assets.

\paragraph{Initial Capital Allocation.}
To reflect realistic trading scenarios representative of typical investors, we initialize the simulated trading accounts with a fixed capital threshold. Specifically, the starting capital is set to 1,000,000 in the respective local currency (i.e., \$1,000,000 for the U.S. market and ¥1,000,000 RMB for the A-share market). Unlike an infinite-capital assumption, this realistic constraint forces the models to make strategic portfolio allocation and liquidity management decisions when executing their buy and sell signals.

\subsection{Research Questions}

\textbf{RQ1 (The Reasoning Dichotomy):} Does an LLM's intrinsic reasoning architecture dictate its reliance on external strategic guardrails versus autonomous intuition?

\textbf{RQ2 (Regime Dependence):} Is the utility of strict strategic alignment and risk anchoring fundamentally contingent upon prevailing market trends?

\textbf{RQ3 (The Scaling Paradox):} Does strategic execution fidelity scale monotonically with model size, or does an optimal mid-scale parameter region exist?

\textbf{RQ4 (Behavioral Biases):} Do standard LLMs inherently manifest retail-like biases, such as the Disposition Effect, and can strict alignment mitigate them?

\subsection{Main Results}

In this section, we provide a comprehensive analysis of the experimental results. By comparing the performance of models across different architectures (Reasoning vs. Standard), strategic constraints, model scales, and temporal horizons, we identify a series of key behavioral patterns that define the efficacy of LLM-based financial agents.

\begin{table*}[htbp]
\centering
\caption{Performance Comparison of Strategy Modes: Proprietary vs. Open-Source Models in A-shares and US Stocks. To maintain focus, this table presents the most representative variants of each model series. The best and second-best results within each metric are highlighted in \textbf{bold} (with orange shading) and \textit{italics} (with blue shading), respectively.}
\label{tab:strategy_market_comparison}
\resizebox{\textwidth}{!}{
\begin{tabular}{ll cccccc cccccc}
\toprule
\multirow{2}{*}{\textbf{Model}} & \multirow{2}{*}{\textbf{Strategy}} & \multicolumn{6}{c}{\textbf{A-shares}} & \multicolumn{6}{c}{\textbf{US Stocks}} \\
\cmidrule(lr){3-8} \cmidrule(lr){9-14}
& & \textbf{TR(\%)} & \textbf{SR} & \textbf{MDD(\%)} & \textbf{Vol(\%)} & \textbf{WR(\%)} & $\boldsymbol{\alpha}$\textbf{(\%)} & \textbf{TR(\%)} & \textbf{SR} & \textbf{MDD(\%)} & \textbf{Vol(\%)} & \textbf{WR(\%)} & $\boldsymbol{\alpha}$\textbf{(\%)} \\
\midrule
\multicolumn{14}{c}{\textbf{Panel A: Proprietary State-of-the-Art Models}} \\
\midrule
GPT-5.4 & Strict & 12.31 & 2.19 & \cellcolor{secondblue}\textit{3.17} & 15.95 & 54.55 & 17.88 & \cellcolor{bestorange}\textbf{$-$0.40} & $-$0.04 & 11.66 & 20.88 & 25.81 & 6.72 \\
GPT-5.4 & Guided & 6.36 & 1.15 & 4.76 & 15.87 & 57.58 & $-$1.87 & $-$10.50 & $-$1.31 & 20.83 & 23.22 & 21.62 & $-$21.97 \\
Claude-Sonnet-4.5 & Guided & 11.52 & 1.81 & 4.76 & 18.36 & 52.38 & 6.68 & $-$4.33 & $-$0.44 & 18.13 & 25.16 & 44.83 & $-$1.64 \\
Gemini-3.1-Pro & Strict & 0.60 & 0.12 & 7.86 & 25.10 & 46.15 & $-$32.53 & $-$8.60 & $-$0.51 & 29.03 & 38.68 & 33.33 & $-$6.66 \\
Gemini-3.1-Pro & Free & 6.38 & 0.80 & 5.88 & 25.43 & 41.18 & $-$17.64 & $-$14.73 & $-$0.56 & 29.22 & 37.38 & 36.84 & $-$8.14 \\
\midrule
\multicolumn{14}{c}{\textbf{Panel B: Open-Source Frontier Models}} \\
\midrule
Qwen3.5-122B-A10B & Guided & \cellcolor{bestorange}\textbf{13.68} & 1.62 & 6.93 & 25.06 & 70.00 & 1.84 & $-$9.19 & $-$0.82 & 25.01 & 29.95 & 35.48 & $-$13.63 \\
Qwen3.5-35B-A3B & Strict & \cellcolor{secondblue}\textit{12.33} & 1.36 & 5.84 & 27.61 & 56.25 & $-$5.74 & $-$13.67 & $-$0.48 & 29.24 & 38.17 & 20.00 & $-$4.86 \\
DeepSeek-Reasoner & Free & 7.62 & 1.57 & \cellcolor{bestorange}\textbf{1.98} & 21.14 & 33.33 & 1.77 & $-$5.35 & $-$0.35 & 22.65 & 33.63 & 34.29 & 0.28 \\
DeepSeek-Chat & Strict & 7.04 & 1.09 & 5.40 & 19.13 & 40.00 & $-$5.79 & $-$3.62 & $-$0.23 & 19.48 & 31.61 & 31.43 & 4.05 \\
GLM-5\_think & Free & 7.82 & 1.23 & 5.33 & 18.64 & 61.76 & $-$3.27 & $-$12.01 & $-$0.35 & 20.76 & 31.77 & 20.83 & 0.52 \\
GLM-5\_nothink & Strict & 6.32 & 1.24 & 4.73 & 14.42 & 44.12 & $-$2.89 & \cellcolor{secondblue}\textit{$-$0.51} & $-$0.01 & 16.81 & 24.41 & 50.00 & 8.83 \\
Kimi-K2.5\_think & Free & 11.68 & 1.40 & 6.79 & 25.13 & 75.00 & $-$3.86 & \cellcolor{secondblue}\textit{$-$0.51} & $-$0.34 & \cellcolor{bestorange}\textbf{1.61} & 42.18 & 33.33 & 0.53 \\
Kimi-K2.5\_nothink & Guided & 12.05 & 1.45 & 7.19 & 24.92 & 65.22 & $-$2.50 & $-$1.03 & 0.02 & 19.55 & 32.65 & 42.86 & 12.02 \\
Minimax-M2.5 & Guided & 10.55 & 1.70 & 4.67 & 17.54 & 66.67 & 5.23 & $-$4.61 & $-$0.22 & \cellcolor{secondblue}\textit{10.28} & 28.74 & 30.77 & 4.36 \\
\bottomrule
\end{tabular}
}
\end{table*}

\subsubsection{Market-Regime Dependent Utility of Strict Alignment}

Our results demonstrate that the utility of the \textit{Strat-LLM} modes is highly sensitive to market regimes. In the uptrending A-share sample, the flexibility of \textit{Free} and \textit{Guided} modes allows models to capture momentum. However, in the downtrending US stock environment, the \textit{Strict Mode} provides a critical safety net. 

As evidenced in Panel A of Table \ref{tab:strategy_market_comparison}, the proprietary model \textbf{GPT-5.4} drastically reduces its Maximum Drawdown (MDD) from 20.83\% (Guided) to 11.66\% (Strict), effectively preserving capital in a declining market. This phenomenon is similarly reflected in open-source models, confirming that \textit{Strict Alignment} is not merely a performance-enhancing tool, but a universal, regime-dependent insurance mechanism that mitigates extreme downside risks.

\subsubsection{The Reasoning Dichotomy: Internal Logic vs. External Guardrails}

Beyond market regimes, our analysis reveals a fundamental dichotomy in how architectural paradigms interact with the \textit{Strat-LLM} framework. As shown in the ablation study (Table \ref{tab:think_vs_nothink}), the necessity of strategic constraints is inversely proportional to the model's intrinsic reasoning capacity (CoT).

\paragraph{Reasoning Models (Think/Reasoner):} Models with native reasoning chains (e.g., Kimi-K2.5\_think, DeepSeek-Reasoner, GLM-5\_think) consistently achieve peak performance in \textit{Free Mode}. For instance, returning to Table \ref{tab:strategy_market_comparison}, Kimi-K2.5\_think reaches an impressive TR (11.68\%) and Win Rate (75.00\%) when unconstrained. For these architectures, external strict rules often clash with their internal multi-step deduction, leading to suboptimal trade timing.

\paragraph{Standard Models (Chat/Non-think):} In contrast, standard conversational models exhibit high vulnerability to market noise when granted full autonomy. Our ablation data (Table \ref{tab:think_vs_nothink}) shows that standard models frequently experience a collapse in performance without rules (e.g., DeepSeek-Chat achieving its optimal 7.04\% TR under \textit{Strict Mode}). This confirms that for architectures lacking deep sequential reasoning, \textit{Strict Mode} acts as an essential ``Risk Anchor,'' preventing logical divergence in volatile environments.

\begin{table}[htbp]
\centering
\caption{Ablation Study: The Impact of Reasoning (Think) on Strategic Preference}
\label{tab:think_vs_nothink}
\resizebox{\columnwidth}{!}{
\begin{tabular}{llccc}
\toprule
\textbf{Model Architecture} & \textbf{Strategy} & \textbf{TR(\%)} & \textbf{SR} & \textbf{MDD(\%)} \\
\midrule
\multirow{3}{*}{\textbf{Kimi-K2.5\_think}} & \textbf{Free} & \textbf{11.68} & 1.40 & 6.79 \\
& Guided & 7.71 & 1.00 & 9.53 \\
& Strict & 9.10 & 1.10 & 8.09 \\
\midrule
\multirow{3}{*}{\textbf{Kimi-K2.5\_nothink}} & Strict & 10.93 & 1.41 & 7.36 \\
& \textbf{Guided} & \textbf{12.05} & 1.45 & 7.19 \\
& Free & 9.15 & 1.29 & 5.75 \\
\midrule
\multirow{3}{*}{\textbf{DeepSeek-Reasoner}} & \textbf{Free} & \textbf{7.62} & 1.57 & 1.98 \\
& Guided & 5.13 & 1.57 & 2.26 \\
& Strict & 3.52 & 0.55 & 6.06 \\
\bottomrule
\end{tabular}
}
\end{table}

\subsubsection{The Scaling Paradox: Mid-Scale Sweet Spot in Financial Alignment}

\begin{figure}[H]
    \centering
    \includegraphics[width=0.85\linewidth,trim=0 20 0 20,clip]{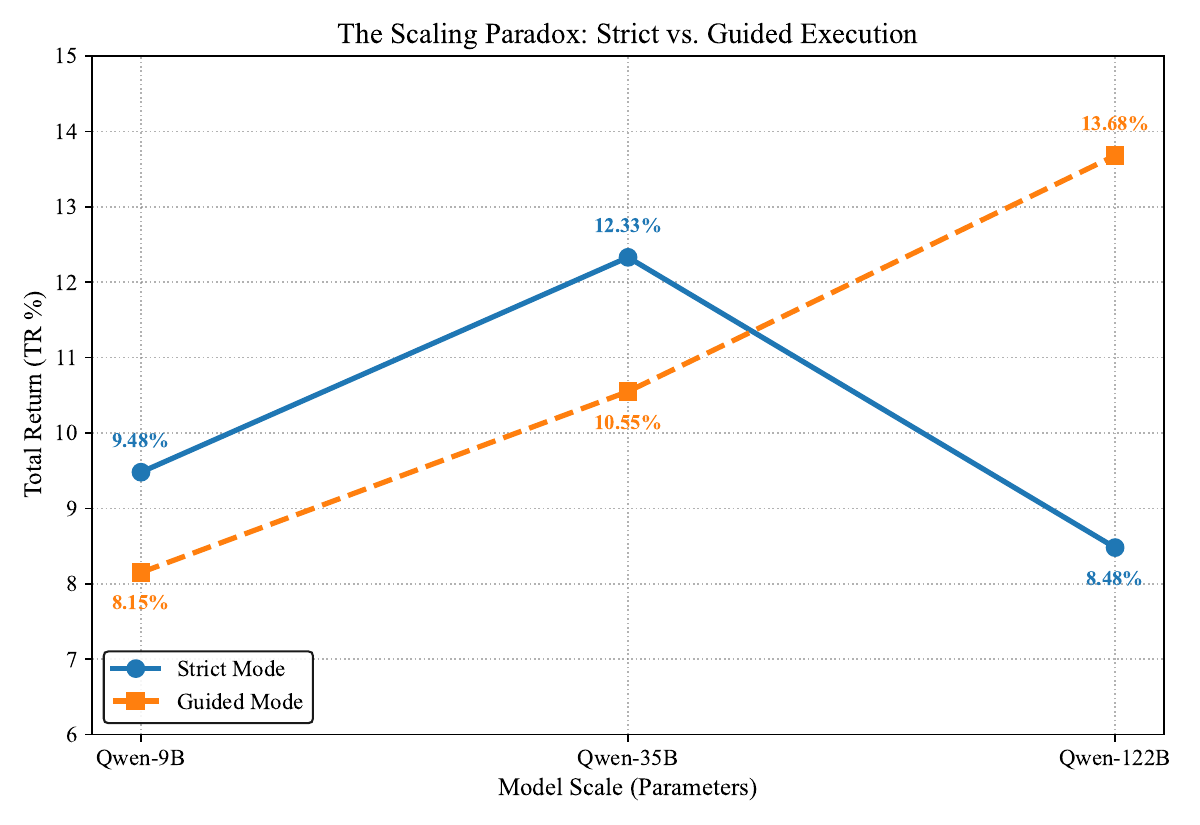}
    \caption{The Scaling Paradox: Total Return comparison across Qwen variants (9B, 35B, 122B). The mid-scale 35B model exhibits optimal execution fidelity under strict constraints.}
    \label{fig:scaling_curve}
\end{figure}

A pivotal finding of our study is the non-linear relationship between model scale and alignment efficacy within the open-source array. While Table \ref{tab:strategy_market_comparison} presents the top-performing representatives, Figure \ref{fig:scaling_curve} illustrates the full ablation across the Qwen series, revealing a distinct ``Mid-Scale Sweet Spot'' for strategic execution.
Specifically, the \textbf{Qwen3.5-35B-A3B} model achieves an exceptional Total Return (TR) of 12.33\% under \textit{Strict Mode}, maintaining robust performance in constrained settings. We hypothesize that mid-scale models possess sufficient knowledge to interpret financial signals while exhibiting high fidelity to instruction-following, avoiding the ``reasoning inertia'' often found in ultra-large models. Conversely, the ultra-large \textbf{Qwen3.5-122B-A10B} excels under \textit{Guided Mode} (13.68\% TR), suggesting that larger models require a degree of strategic autonomy to leverage their complex internal heuristics. This indicates that the ``Alignment Tax'' is highest for ultra-large models when subjected to rigid rule-sets, but transforms into an ``Alignment Premium'' when shifted to guided frameworks.

\subsubsection{The High Win-Rate Trap: Emergent Disposition Effect}

In quantitative trading, a high Win Rate (WR) is frequently misconstrued as a definitive proxy for strategy robustness. However, our empirical analysis reveals a counterintuitive asymmetry between WR and Total Return (TR) / Alpha ($\alpha$), exposing a critical vulnerability in models that lack deep reasoning or strict alignment constraints. 

As evidenced in Table \ref{tab:strategy_market_comparison}, higher win frequencies do not linearly translate to superior portfolio growth. In the A-share environment, \textbf{Minimax-M2.5} (Guided) achieves an inflated WR of 66.67\% but yields a TR of 10.55\% and an $\alpha$ of 5.23\%. Conversely, the proprietary \textbf{GPT-5.4} (Strict) secures a dominant TR of 12.31\% and an unparalleled $\alpha$ of 17.88\%, despite a significantly lower WR of 54.55\%. This stark divergence is even more pronounced in the US bear market: \textbf{GLM-5\_nothink} (Strict) boasts a 50.00\% WR yet remains in negative territory (TR $-$0.51\%), whereas \textbf{GPT-5.4} (Strict) dictates the best bear-market performance (TR $-$0.40\%, $\alpha$ 6.72\%) with a mere 25.81\% WR.

Viewed through behavioral finance, standard LLMs inherently exhibit the \textit{Disposition Effect} by prematurely securing small gains while tolerating massive drawdowns. In contrast, advanced reasoning architectures act as cognitive debiasing mechanisms that prioritize mathematical expectancy and risk-reward ratios over inflated win rates. Consequently, overcoming these emergent retail-investor biases in LLM-driven agents necessitates either intrinsic Chain-of-Thought processing or extrinsic \textit{Strict Mode} constraints.

\subsubsection{Temporal Horizon Sensitivity}

A consistent performance gap is observed across temporal scales. The \textbf{long-term strategic window} (90 trading days) significantly outperforms the \textit{short-term tactical window} (15 trading days). Over short horizons, models struggle with signal ambiguity, resulting in indecisive trading behavior. In contrast, extended horizons provide sufficient latency for the models’ entry logic to materialize, supporting the view that LLM-based reasoning is better suited for strategic positioning rather than high-frequency tactical adjustments.

\section{Conclusion}
In this study, we introduced \textit{Strat-LLM}, a framework that shifts financial agent evaluation from static profitability to \textbf{stratified strategic auditability}. Extensive stress tests reveal that the optimal level of strategic constraint depends heavily on model architecture and market conditions. Specifically, reasoning-heavy models thrive under full autonomy, whereas standard models require rigid alignment to mitigate downside risks. Furthermore, we observe a \textbf{mid-scale advantage} at approximately 35B parameters and find that strict rules penalize ultra-large models unless they are granted guided autonomy, challenging conventional scaling laws in financial tasks. We also expose an emergent disposition effect where models misleadingly optimize for high win rates over total returns, a behavioral bias that can only be mitigated through deep reasoning or strict external guardrails. Ultimately, \textit{Strat-LLM} provides a rigorous baseline for trustworthy autonomous trading, bridging the gap between unregulated generation and professional compliance.

\bibliographystyle{IEEEtran} 
\bibliography{references}    
\end{document}